\documentclass[10pt,twocolumn,letterpaper]{article}

\usepackage{iccv}
\usepackage{times}
\usepackage{epsfig}
\usepackage{graphicx}
\usepackage{amsmath}
\usepackage{amssymb}

\usepackage{bbding}
\usepackage{soul}
\usepackage{bibunits}
\usepackage{bm}
\usepackage{array}
\usepackage[table]{xcolor}
\usepackage{pdflscape}
\usepackage{makecell}
\usepackage{framed,multirow}
\usepackage[flushleft]{threeparttable}
\usepackage{amssymb}% http://ctan.org/pkg/amssymb
\usepackage{pifont}% http://ctan.org/pkg/pifont
\usepackage{algorithm}
\usepackage{listings}
\usepackage{pythonhighlight}

\usepackage{amsmath}
\DeclareMathOperator*{\argmax}{arg\,max}

\usepackage[pagebackref=true,breaklinks=true,letterpaper=true,colorlinks,bookmarks=false]{hyperref}

\newcommand{\methodname}{\mbox{OccNeRF}}

\definecolor{Highlight}{HTML}{39b54a}  % green

\newcolumntype{P}[1]{>{\centering\arraybackslash}p{#1}}
\newlength\savewidth

\newcolumntype{x}[1]{>{\centering\arraybackslash}p{#1pt}}
\newcolumntype{z}[1]{>{\raggedright\arraybackslash}p{#1pt}}

\definecolor{best_color}{HTML}{FCE5CD}
\definecolor{better_color}{HTML}{DEEDF2}

\iccvfinalcopy % *** Uncomment this line for the final submission

\def\iccvPaperID{6487} % *** Enter the ICCV Paper ID here

\ificcvfinal\fi

\begin{document}

%%%%%%%%% TITLE

\title{\vspace{-1em}Rendering Humans from Object-Occluded Monocular Videos}

\author{Tiange Xiang\thanks{Correspondence to \href{mailto:xtiange@stanford.edu}{xtiange@stanford.edu}}, Adam Sun, Jiajun Wu, Ehsan Adeli, Li Fei-Fei\\
Stanford University\\
% \and
}

\maketitle
\ificcvfinal\thispagestyle{empty}\fi

% (or \textbf{Occ-HNeRF} or \textbf{MOON}?? short for render \textbf{M}oving human in \textbf{O}bject-\textbf{O}ccluded videos by \textbf{N}erf)

%%%%%%%%% ABSTRACT
\begin{abstract}
    3D understanding and rendering of moving humans from monocular videos is a challenging task. Despite recent progress, the task remains difficult in real-world scenarios, where obstacles may block the camera view and cause partial occlusions in the captured videos. Existing methods cannot handle such defects due to two reasons. First, the standard rendering strategy relies on point-point mapping, which could lead to dramatic disparities between the visible and occluded areas of the body. Second, the naive direct regression approach does not consider any feasibility criteria (\ie, prior information) for rendering under occlusions. To tackle the above drawbacks, we present \textbf{\methodname}, a neural rendering method that achieves better rendering of humans in severely occluded scenes. As direct solutions to the two drawbacks, we propose surface-based rendering by integrating geometry and visibility priors. We validate our method on both simulated and real-world occlusions and demonstrate our method's superiority. Project page: \url{https://cs.stanford.edu/~xtiange/projects/occnerf/}
\end{abstract}

%%%%%%%%% BODY TEXT
\section{Introduction}
Rendering 3D human bodies from a sequence of observations is of great interest in various communities, including robotics~\cite{yang2021s3}, motion analysis~\cite{endo2022gaitforemer}, and healthcare~\cite{gerats20223d}. This task is challenging, since one must recover the complete human body with complex textures and poses from sparse partial observations. It is usually cumbersome to acquire images of the same human object from multiple camera angles simultaneously; hence, capturing a monocular video from a single camera is more common and feasible. 

\begin{figure}[t]
    \centering
    \includegraphics[width=\linewidth]{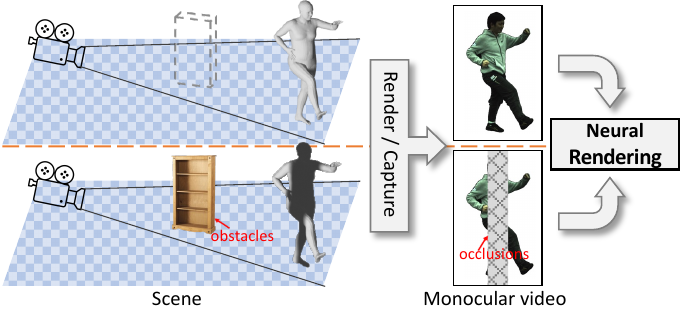}
    \caption{Object obstacles in the scene may cause severe occlusions in the rendered/captured videos, imposing additional challenges into the rendering process. \textbf{Top row:} Ideal scene with no defects and clear view of the body; \textbf{Bottom row:} Real-world scene with undesired obstacles and occluded body parts.}
    \vspace{-1em}
    \label{fig:intro}
\end{figure}

The task of rendering humans from a monocular video is not new. Progress so far mainly focuses on rendering quality \cite{neuralbody, weng2022humannerf} and rendering efficiency \cite{peng2022selfnerf, jiang2022instantavatar}. However, most existing neural rendering methods assume that the human object is placed in a scene with a clear view of the entire body without any external interference. In contrast, real-world environments often contain undesired obstacles that contaminate training data and impact the rendering quality (See \figureautorefname~\ref{fig:intro}). These real-world occlusions pose significant challenges for training when using only monocular videos, where no other camera angles can be used to provide complementary information. As a result, a direct application of previous neural rendering methods on object-occluded videos leads to subpar performance. Optimizing a neural radiance field is difficult under occlusions. There is often no ground truth associated with the occluded area. Additionally, radiance fields are typically optimized in a scene-specific manner; that is, no external information can and should be used to fill in the occluded areas.

Two major drawbacks of previous methods impair their capabilities to train on object-occluded videos. First, the prior work does not account for local geometry cues in their rendering process. Following the point-based rendering paradigm as in NeRF \cite{nerf}, most previous methods render color and density values of a ray sample by only looking at a single 3D coordinate. However, we explain in \sectionautorefname~\ref{sec:3.2} that this basic strategy may lead to dramatically different rendering results even in very close positions. Second, methods suffer from not properly incorporating priors. In the monocular video setting, geometry (\eg, SMPL \cite{SMPL}) and visibility priors can describe a complete human geometry and indicate which body parts are visible to the camera. 

In this work, we propose novel methods for dealing with the above drawbacks, allowing us to accurately render occluded humans from monocular video. We first present a surface-based rendering strategy that determines the radiance of each 3D ray sample by conditioning it on a wide region of the human body's surface. A geometry prior is used to discretely parameterize the surface segments. We then collect visibility frequencies on the human body through training frames and formulate them as attention maps for better aggregation of the surface regions. Finally, we design a loss function to encourage the network to output high-density values for positions within the human body.

In summary, our contributions are three-fold: \textbf{\emph{(i)}} We are the first to study dynamic human rendering under real-world settings with severe occlusions. \textbf{\emph{(ii)}} We propose novel methods that include surface-based rendering, a reformulation of body part visibility frequency as attention, and a completeness loss to enable human rendering from object-occluded monocular videos. \textbf{\emph{(iii)}} We empirically demonstrate that our methods achieve significant quantitative and qualitative improvements compared to the previous state-of-the-art, yielding the first baseline in this topic.

\section{Related Work}

\noindent\textbf{3D Human Modeling.} 
Reconstructing the appearance and geometry of humans has always been challenging. From \cite{matusik2000image, collet2015high, dou2016fusion4d}, techniques have been consistently designed for high-quality and efficient human modeling. Traditional methods mainly relied on SMPLify \cite{Bogo:ECCV:2016} or Video-avatars \cite{alldieck2018video} to regress SMPL \cite{SMPL} to parameterize a structured human body. More complex networks were subsequently designed that can model 3D humans based on temporal priors \cite{humanMotionKZFM19, kocabas2019vibe}, based on depth \cite{jiang2022selfrecon, saito2019pifu, hong2021stereopifu}, or multiple human instances simultaneously \cite{jiang2020coherent, sun2021monocular, BEV, zhang2021body}. Although this line of methods can generate a reasonable human body mesh fast, using parametric SMPL models limit their ability to achieve photo-realistic view synthesis.

%their fast generation of the rough human body surface can be used as good geometry priors. 

% However, recent advances in neural representations have made it possible to directly reconstruct high-fidelity 3D human avatars from sparse views or a single video without the need for a personalized template. Despite their impressive quality, these methods can suffer from slow rendering and training speeds due to the canonical representation's slow speed and deformation algorithms. To address this issue, the authors propose a method that enables learning of avatars within minutes.

% Other methods use traditional model-based techniques, such as SMPLify and Videoavatars, to obtain per-frame parameters via optimization, while others use networks to learn the priors of humans from ground truth data. These methods, such as Kolotouros et al.'s GraphCNN-based approach and PIFu, PIFuHD, StereoPIFu, and BCNet, employ various inputs and techniques to improve the accuracy of the reconstructed models' geometry.

\noindent\textbf{Neural Radiance Field for Human Rendering.}
Since the emergence of Neural Radiance Fields (NeRF) \cite{nerf}, different extensions have been recently developed to enable high-quality rendering of static scenes \cite{hedman2021baking, srinivasan2021nerv, barron2021mip, barron2022mip, verbin2022ref, tancik2020fourier, sun2021direct, muller2022instant}, moving objects \cite{gao2021dynamic, li2021neural, park2021nerfies, park2021hypernerf, pumarola2021d, ost2021neural}, and dynamic humans \cite{neuralbody, bergman2022generative, chen2021animatable, chibane2020implicit, chen2022gdna, corona2021smplicit, deng2020nasa, feng2022capturing, he2020geo, he2021arch++, huang2020arch, jiang2022selfrecon, li2022tava, liu2021neural, noguchi2021neural, peng2021animatable, tiwari2021neural, wang2022arah, xu2021h, peng2022selfnerf, jiang2022instantavatar, jiang2022neuman}. NeRF predicts the color and density of each ray sample point in a 3D space and aggregates them together through volume rendering (more details are in \sectionautorefname~\ref{sec:preliminary}). This approach enables the capture of intricate lighting effects and textural details that are typically difficult to model in traditional methods.

Our work is built upon HumanNeRF \cite{weng2022humannerf} due to its state-of-the-art rendering quality for monocular videos. HumanNeRF maintains a static T-pose human body as the canonical space and learns a motion field \cite{weng2020vid2actor} that maps the canonical representation to every frame of the video in the observation space (more details are in \sectionautorefname~\ref{sec:preliminary}). We note that a concurrent work, SelfNeRF \cite{peng2022selfnerf}, shares a similar regression schema as ours. However, their method is designed particularly for fast rendering and compromises rendering quality. Moreover, all of the above approaches were developed on clean training data only, where body parts are assumed to be clearly demonstrated in the monocular video without any occlusions. On the other hand, our work aims to render humans under occlusions.

\noindent\textbf{Occluded Human Modelling.} Rendering objects under all kinds of real-world defects, especially partial occlusions, is a long-standing research problem. Early works sought to estimate human poses from occluded images and videos \cite{sarandi2018robust, zhang2020object, kocabas2021pare, biggs20203d}, while more recent works \cite{rockwell2020full, sun2021monocular, yang2022lasor, kocabas2021pare} learn both SMPL shape and pose priors directly from occluded images and videos. The generated SMPL parameters from these robust methods can be used as good geometric prior for a subsequent rendering process.

However, optimizing a NeRF from occluded images is still an unsolved problem. There are very few works that were specifically designed for rendering under occlusions. NeuRay \cite{liu2022neural} was proposed to regress not only the radiance but also a feature vector of every ray sample to indicate visibility. This enables the optimization of the radiance field to focus on visible features and reduce interference from occlusions. Ha-NeRF \cite{chen2022hallucinated} presents an appearance hallucination module to handle time-varying appearances and an anti-occlusion module to decompose the static subjects for visibility accurately. Unfortunately, these existing methods are not capable of handling dynamic objects, and the multi-view inputs used in past work actually make it easier to learn under occlusions. In this work, we consider visibility as an additional prior to assist in rendering under occlusions. Our work is the very first in this field to handle occlusions for rendering dynamic objects from only a monocular video.

\section{Methods}
\begin{figure*}[t]
    \centering
    \includegraphics[width=\linewidth]{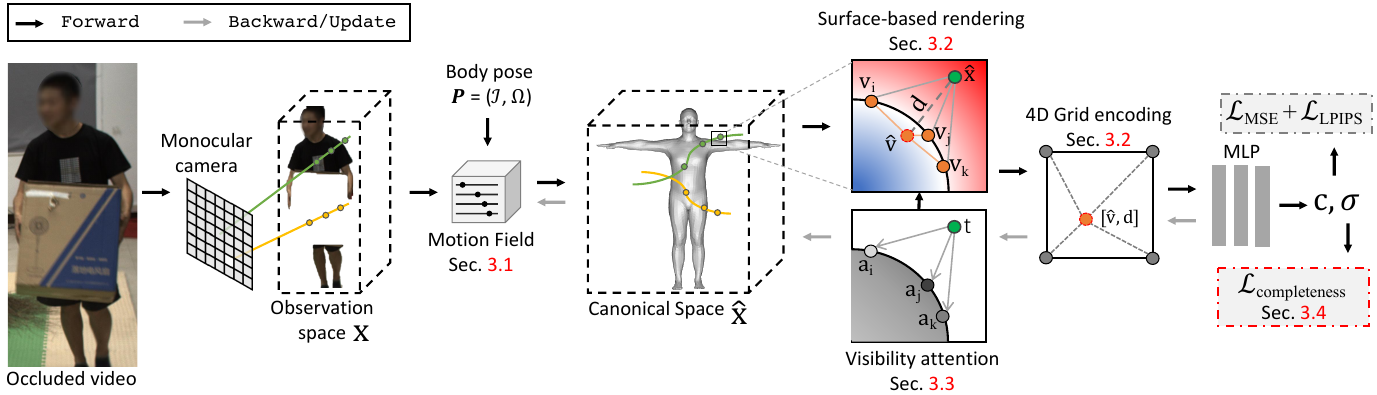}
    \caption{\textbf{\methodname} functions on video frames and optimizes a neural radiance field for synthesizing novel views of an object-occluded human. With a pre-computed body pose, we first adopt the motion field to map observable ray samples $\mathbf{x}$ into coordinates $\mathbf{\hat{x}}$ in a canonical space. Nearest parameterization vertices $\{\mathbf{v}_i\}$ of every $\mathbf{\hat{x}}$ are searched and conditioned by our surface-based rendering method. During training, we iteratively update the attention scores $\{\mathbf{a}\}$ for all $\{\mathbf{v}\}$ as indications of their visibility. This ensures more attention on frequently visible vertices to improve rendering quality. The blended vertex $\mathbf{\hat{v}}$ along with its signed distance to $\mathbf{\hat{x}}$ are jointly encoded via a 4D hash grid before being fed into the regression MLP along with the encoded vertices. Photometric and perceptual constraints are enforced against visible pixels, while an additional loss function is designed to encourage geometry completeness in occluded areas.}
    \label{fig:framework} 
    \vspace{-1em}
\end{figure*}

In this section, we first review preliminaries and background in NeRF \cite{nerf} and HumanNeRF \cite{weng2022humannerf} (\sectionautorefname~\ref{sec:preliminary}). We then present our \methodname\ by introducing a new rendering strategy (\sectionautorefname~\ref{sec:3.2}), a formulation of visibility into attention (\sectionautorefname~\ref{sec:3.3}), and a novel loss function (\sectionautorefname~\ref{sec:3.4}) to ensure high rendering quality as well as geometry completeness under occlusions. An overview of our \methodname\ is shown in \figureautorefname~\ref{fig:framework}.

\subsection{Preliminaries and Background} \label{sec:preliminary}

\noindent\textbf{Neural Radiance Field \cite{nerf}.} Consider a (bounded) 3D scene. NeRF learns a regression function $\mathcal{F}$ (usually an MLP) that takes the encoded coordinates of a 3D point $\mathbf{x}\in\mathbb{R}^{3}$ in the scene as input, and outputs the corresponding color $\mathbf{c}$ and density $\mathbf{\sigma}$ at that position:
\begin{equation} \label{eq:point}
    \mathbf{c}, \mathbf{\sigma} = \mathcal{F}(\gamma(\mathbf{x})),
\end{equation}
where $\gamma(\cdot)$ is an encoding function. We refer to the above point-to-point mapping as \textbf{point-based rendering}. Instead of sampling points $\mathbf{x}$ randomly in the scene, NeRF casts rays $\mathbf{r}$ towards the directions $\pi$ from the camera origin $\mathbf{o}$ to every pixel, and sample $\mathbf{x}$ on the rays uniformly. Then, NeRF renders the pixel by aggregating the regressed color and density at each $\mathbf{x}$ via volume rendering \cite{lombardi2019neural}:
\begin{equation}
    \sum_{i} \alpha(\mathbf{x}_i)\prod_{j<i}(1-\alpha(\mathbf{x}_j))\mathbf{c},
\end{equation}
where $\alpha(\mathbf{x}_i)=1-\exp(-\mathbf{\sigma}_i\delta_i)$, $\mathbf{x}_i=\mathbf{o}+z_i\pi$, $z_i$ is the z-axis position of ray samples, and $\delta_i=z_{i+1}-z_{i}$ is the distance between two samples along the ray.

\noindent\textbf{HumanNeRF \cite{weng2022humannerf}.} HumanNeRF is a method based on NeRF that can render humans from monocular videos by representing them as neural fields. The method first defines a moving human in a static canonical space with 3D coordinates $\hat{\mathbf{x}}$, and warps the human in different dynamic poses by warping the canonical body pose $\mathbf{p}$ to the observation space. This warping process also defines the transformation of 3D coordinates in the two spaces:
\begin{equation}
    \hat{\mathbf{x}} = \mathcal{T}(\mathbf{p}, \mathbf{x}),
\end{equation}
where $\mathcal{T}$ is a network that maps $\mathbf{x}$ in the observation space to corresponding coordinates $\hat{\mathbf{x}}$ at the canonical space, denoted as the \textbf{motion field}. The motion field  achieves the mapping by performing a weighted sum of a set of $K$ motion bases defined by rotations $R_i$ and translation $t_i$ of the $i_{th}$ bone of the human body:
\begin{equation}
    \hat{\mathbf{x}} = \sum^{K}_{i}w_i(\mathbf{x})(R_i\mathbf{x}+t_i),
\end{equation}
where $R_i$ and $t_i$ can be directly computed from $\mathbf{p}$. $w_i$ serves as the weights in the observation space, which can be approximated using the weights defined in the canonical space. Similar to \cite{weng2023personnerf}, we removed both the non-rigid motion and the pose correction part of the motion field.

\subsection{Surface-based Rendering} \label{sec:3.2}
\noindent\textbf{Motivation.} Although HumanNeRF and its variants can already achieve decent rendering quality in an occlusion-free scene, they fail to excel when obstacles block the view of the camera that causes severe occlusions. We attribute this failure to the point-based rendering strategy (reviewed in \sectionautorefname~\ref{sec:preliminary}). Given the ray samples at discrete 3D coordinates $\mathbf{x}$ in a continuous 3D space, even a mild variation between two coordinates $\mathbf{x}_a$ and $\mathbf{x}_b$ can lead to dramatic disparities on the outputs. Let there be no overlaps between the input distributions $\{\mathbf{x}_a\}$ and $\{\mathbf{x}_b\}$:
\begin{equation}
    \{\mathbf{x}_a\} \cap  \{\mathbf{x}_b\} = \emptyset\ \mathbin\vert\ \mathbf{x}_a\not\equiv\mathbf{x}_b.
\end{equation}
Then, in an occluded scene, when only $\mathbf{x}_a$ is visible to the camera, non-overlapping inputs may yield huge output differences, even at very close locations. This is because $\mathbf{x}_a$ has visible supervisions while $\mathbf{x}_b$ does not, which leads to unexpected artifacts and unstable rendering quality at occluded regions.

\begin{figure}[t]
    \centering
    \includegraphics[width=\linewidth]{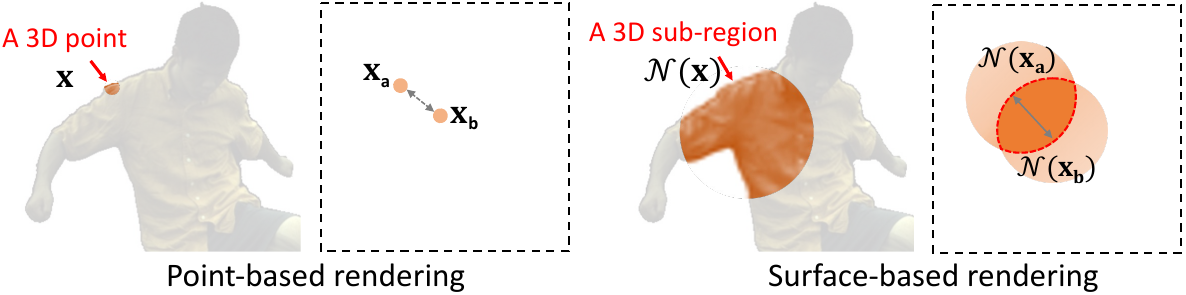}
    \caption{\textbf{Left:} Point-based rendering takes as input a single 3D point $\mathbf{x}$ that has no overlap with nearby (but not identical) points, and is poorly conditioned at occluded areas; \textbf{Right:} Our surface-based rendering approach takes as input a 3D sub-regions $\mathcal{N}(\mathbf{x})$ at location $\mathbf{x}$ that yields a large overlap at adjacent locations for better conditioning at occluded areas.}
    \label{fig:2}
    \vspace{-1em}
\end{figure}

This motivates us to enlarge the range of the inputs to cover a wider range of 3D space rather than a single 3D coordinate. We expect that a new rendering strategy with range-to-point mapping will be able to reduce the output difference at adjacent locations:
\begin{equation}
   \int_{\mathbb{R}^3}\min[\mathcal{N}(\mathbf{x}_a), \mathcal{N}(\mathbf{x}_b)]d\mathbf{x} \gg 0,
\end{equation}
where $\mathcal{N}(\mathbf{x}_a)$ and $\mathcal{N}(\mathbf{x}_b)$ are 3D sub-regions corresponding to the target coordinates $\mathbf{x}_a$ and $\mathbf{x}_b$. With a focus on human rendering, we define the sub-regions as continuous segments on the body surface. We name this rendering strategy \textbf{surface-based rendering}. A high-level comparison to the standard point-based rendering is outlined in \figureautorefname~\ref{fig:2}.

% \begin{equation}
%    |\{\mathbf{x}_a^{(0)}\cdots\mathbf{x}_a^{(N)}\} \cap  \{\mathbf{x}_b^{(0)}\cdots\mathbf{x}_b^{(N)}\}| \gg |\emptyset| =0,
% \end{equation}
% where $\mathbf{x}_a^{(i)}$ and $\mathbf{x}_b^{(i)}$ are the parameterized coordinates that are used to regress $\mathbf{c}_a, \mathbf{\sigma}_a$ and $\mathbf{c}_b, \mathbf{\sigma}_b$, respectively.

%the difference between $|\mathcal{F}(\gamma(\mathbf{x}_a)) - \mathcal{F}(\gamma(\mathbf{x}_b))|$ 

% To regress color and density at each ray sample, we first need to define a corresponding 3D sub-space that spans a bigger set of coordinates. Since we focus on human rendering in this work, we define the sub-space as continuous segments on the surface of the human body. We call this rendering process \textbf{surface-based rendering}.

% \footnote{As mentioned in \cite{peng2022selfnerf}, a more accurate geometry prior \cite{jiang2022selfrecon} can be used for better results. We used SMPL in this work for simplicity.}

\noindent\textbf{Parameterization.}
It is difficult to process continuous variables, especially ones with irregular distributions, which is the case with human surfaces. We approach this challenge by using a discretized parameterization of the continuous 3D sub-regions. Specifically, we use the pre-computed SMPL \cite{SMPL} mesh as a geometry prior to roughly outline the surface of the human body. The surface segments are then parameterized by the $k$ nearest mesh vertices $\{\mathbf{v}_1,\cdots,\mathbf{v}_k\}$ when using the target coordinates $\mathbf{x}$ as queries. We denote these discrete neighboring coordinates as parameterization vertices. With our surface-based rendering approach, we now reformulate \equationautorefname~\ref{eq:point} as a hybrid combination of both the target coordinate and the parameterized surface:
\begin{equation} \label{eq:aggregation}
    \mathbf{c}, \mathbf{\sigma} = \mathcal{F}(\underbrace{\gamma(\mathbf{\hat{x}})}_{\text{point term}} \mathbin\Vert\ \ \underbrace{\phi(\{\gamma(\mathbf{v}_1), \cdots, \gamma(\mathbf{v}_k)\})}_{\text{surface term}}\ ),
\end{equation}
where $\phi$ is a function that aggregates all $\{\mathbf{v}_i\}$ of a query $\mathbf{x}$ and $\mathbin\Vert$ denotes concatenation. The above formulation requires all parameterization points $\mathbf{v}$ to be as accurately laid on the human body as possible. However, this is difficult for the coarsely structured SMPL mesh with potential approximation errors. Therefore, we rely on the inaccurate SMPL mesh only as an initialization and enable the positions of $\mathbf{v}$ to be optimized jointly with the network. This formulation is analogous to area sampling \cite{Loubet2019Reparameterizing} for ray tracing, which not only integrates samples along the ray but in vicinity area.

\noindent\textbf{Multi-Scale Representations.} Choosing the area of surface segments and the number of parameterization vertices $k$ per query is another issue. A small area leads to less overlap and more unstable results, while a large area leads to more overlap of $\{\mathbf{v}_i\}$ at two query locations but a more inefficient search of nearest neighbors. Taking inspiration from the multi-scale mechanism used in I-NGP \cite{muller2022instant}, we construct the set of parameterization vertices by finding the nearest neighbors on the SMPL mesh at multiple scales. We define the default SMPL mesh at the finest scale and iteratively down-sample the mesh with sparse vertices through furthest point sampling \cite{qi2017pointnet} with a ratio of 0.25 for 3 iterations. In practice, we set a small $k=5$ at all 4 scales, which enables an efficient span over a large surface area.

\noindent\textbf{Surface-Aware Regression.} The additional operations introduced above inevitably slow down network training. Similar to \cite{peng2022selfnerf, jiang2022instantavatar}, we adopt a hash grid \cite{muller2022instant} in the canonical space as our encoding function $\gamma(\cdot)$ instead of the standard frequency-based positional encoding \cite{nerf}. Furthermore, for better awareness of the human body surface, we represent a query point in the canonical space $\mathbf{\hat{x}}$ by the combination of its closest parameterization vertex $\mathbf{\hat{v}}$ and their signed distance $\mathbf{d}$. For simplicity, we reuse the previously searched $k$ nearest vertices and blend them through normal similarities to form the closest vertex $\mathbf{\hat{v}}$:
\begin{equation}
    \mathbf{\hat{v}} = \frac{\sum_i^{k}|\texttt{cos}(\mathbf{\hat{x}}, \mathbf{v}_i)|\mathbf{v}_i}{\sum_i^{k}|\texttt{cos}(\mathbf{\hat{x}}, \mathbf{v}_i)|},
\end{equation}
where $\texttt{cos}(\mathbf{\hat{x}}, \mathbf{v}_i)$ denotes the cosine similarity between the vector $\mathbf{\hat{x}} - \mathbf{v}_i$ and the normal vector at $\mathbf{v}_i$. After obtaining $\mathbf{\hat{v}}$, we can easily determine $\mathbf{d}$ between $\mathbf{\hat{v}}$ and $\mathbf{\hat{x}}$ via a multiplication with the normal vectors at $\mathbf{v}_i$. Inspired by \cite{peng2022selfnerf}, we then rely on a 4D hash grid to encode the combination $[\mathbf{\hat{v}}, \mathbf{d}]$. Note that our formulation differs from \cite{peng2022selfnerf}, which encodes a 4D feature vector for every nearest neighbor first and then blends the feature vectors afterward. Our implementation encodes every $\mathbf{\hat{x}}$ only once. 

With the above formulation, we can rewrite the point term $\gamma(\mathbf{\hat{x}})$ in \equationautorefname~\ref{eq:aggregation} into $\gamma([\mathbf{\hat{v}}, \mathbf{d}])$. The surface term is formulated with visibility priors, as discussed below.

\subsection{Visibility Attention} \label{sec:3.3}

In occluded videos, some parts of the human body may be more frequently visible by the camera than others. As a result, more supervision is provided for these highly visible parts which makes $\mathcal{F}$ fit on these \emph{visible areas} much better. When conditioning on a wide range of surfaces, we hope to pay more attention to the highly visible vertices than the hardly visible ones. We achieve this through an attentive aggregation of the neighbor vertices $\{\mathbf{v}_i\}$ via the function $\phi$ (\equationautorefname~\ref{eq:aggregation}) based on their visibility frequency. 

\begin{figure}[t]
    \centering
    \includegraphics[width=\linewidth]{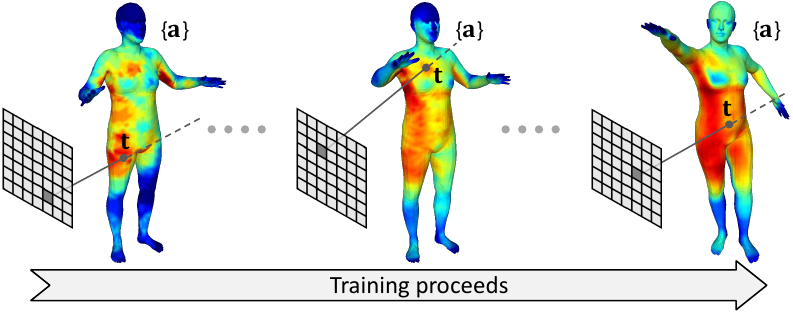}
    \caption{\textbf{Formulating visibility as attention}. Highly visible body parts along with associated parameterization vertices are expected to correspond to more attention.}
    \label{fig:3}
    \vspace{-1em}
\end{figure}

Specifically, for each of the vertices $\mathbf{v}_i$, we maintain a separate attention score $\mathbf{a}_i$ to be updated on-the-fly as the training proceeds. Instead of recording the visibility frequency of all sample points in the camera rays, only the termination point $\mathbf{t}$ per ray should be considered. However, it is computationally expensive to find the exact intersection point between the camera rays and the human body. We approximate $\mathbf{t}$ as the sample point with the highest $\alpha$ along each of the rays, such that $\mathbf{t}=\mathbf{\hat{x}}_{\argmax\{\alpha\}}$. For each $\mathbf{t}$, we again rely on the $k$ nearest vertices $\{\mathbf{v}_i\}$ found earlier to determine the visible area on the body. At each training step, for all neighbors $\{\mathbf{v}_i\}$ of every $\mathbf{t}$, we increment their corresponding attention scores $\{\mathbf{a}_i\}$ by 1. Taking visibility into account, \equationautorefname~\ref{eq:aggregation} can be reformulated as:
\begin{equation} \label{eq:attentive_agg}
    %\phi(\{\gamma(\mathbf{v}_1), \cdots, \gamma(\mathbf{v}_k)\}) = 
    \mathbf{c}, \mathbf{\sigma} = \mathcal{F}(\ \underbrace{\gamma([\mathbf{\hat{v}}, \mathbf{d}])}_{\text{point term}}\ \mathbin\Vert\ \underbrace{\frac{\sum^{k}_i \mathbf{a}_i\gamma([\mathbf{v}_i, \mathbf{d}^{(\mathbf{v}_i)}])}{\sum^{k}_i\mathbf{a}_i}}_{\text{surface term}}\ ),
\end{equation}
where the hash grid encoding $\gamma(\cdot)$ is shared for both point and surface. Recall that all vertices $\mathbf{v}$ have learnable coordinates, and we compute their signed distance $\mathbf{d}^{(\mathbf{v}_i)}$ w.r.t the vertices on the initial SMPL mesh. The updating process of our visibility attention is demonstrated in \figureautorefname~\ref{fig:3}.

\subsection{Loss Functions} \label{sec:3.4}

Following HumanNeRF, we mainly supervise the training of \methodname\ through pixel-wise photometric loss $\mathcal{L}_\text{MSE}$ and LPIPS \cite{zhang2018unreasonable} loss $\mathcal{L}_\text{LPIPS}$ to encourage high-quality rendering at the visible parts. Unfortunately, these constraints do not apply to the occluded parts, where supervisions are hardly available. Hence we design another constraint to explicitly penalize renderings with incomplete geometry and encourage high-density values within the human body. The previously computed signed distances $\mathbf{d}$ are good approximations of the position of ray samples w.r.t the SMPL mesh. Instead of only enforcing the samples near the body surface, we apply the constraint to all samples with negative $\mathbf{d}$. Our completeness loss $\mathcal{L}_\text{comp}$ is therefore defined as:
\begin{equation}
%\begin{split}
     \mathcal{L}_\text{comp} = m\cdot\exp(\texttt{ReLU}(-\texttt{ReLU}(\sigma)+\beta)-\beta),
    %  m &= 
    % \begin{cases}
    % 1, & \text{if } \mathbf{d} < 0,\\
    % 0,              & \text{otherwise},
    % \end{cases}
%\end{split}
\end{equation}
where $m=1$ if $\mathbf{d}<0$ and $0$ otherwise, and $\beta=10$ is a hyper-parameter. Intuitively, it is designed to penalize incompleteness inside the human body. We use \texttt{ReLU} to clip negative $\sigma$ in the range of $[-\beta, 0]$ and use exponential trick to decrease penalty for high densities. \methodname\ is supervised by a weighted combination of the three losses:
\begin{equation} \label{eq:target}
    \lambda_1\mathcal{L}_\text{MSE} +  \lambda_2\mathcal{L}_\text{LPIPS} +  \lambda_3\mathcal{L}_\text{comp}.
\end{equation}

\section{Experiments}
\subsection{Datasets}
\noindent\textbf{ZJU-MoCap \cite{neuralbody}.} This dataset contains humans performing a wide variety of activities. Following HumanNeRF \cite{weng2022humannerf}, we mainly evaluate our methods on the 6 subjects (377, 386, 387, 392, 393, 394) for direct comparisons. Videos captured by \emph{camera 1} are used as training data, and the other 22 cameras are used for evaluation. Since the MoCap data was captured in a lab environment without the interference of any obstacles, we simulate occlusions to be applied to the training videos. Without losing generality, we simulate the presence of a box-like obstacle right in front of the camera that causes a rectangular area at the center of the frame to be occluded. To do so, we first determine a center point of the valid pixels from video frames, and then mask out 50\% of these pixels (demonstrated in \figureautorefname~\ref{fig:intro}). Our simulated obstacle and the occluded area are not intended to be moving along with the subject. Since there is no obvious horizontal movement of subjects, we further expect that they can move out of the occluded area for a short time and therefore only apply the mask to 80\% of the frames. 

\noindent\textbf{OcMotion \cite{huang2022occluded}.} This dataset contains humans interacting with various objects, subject to real-world occlusions. There are a total of 48 videos, and each video was captured at 6 different camera poses. We evaluated on 2 videos with different extents of occlusions. Specifically, we selected 540 frames from \emph{video 14, camera 4} and 500 frames from \emph{video 11, camera 2} as benchmarks for \textbf{mild} and \textbf{severe} real-world occlusions respectively. For both benchmarks, we use the camera matrices, human body poses, and SMPL parameters provided by the dataset, which were computed by \cite{huang2021dynamic} directly on the occluded videos. We provide more results in supplementary materials. We also show the robustness of our method to inaccurately estimated priors.

\begin{figure*}[t]
    \centering
    \includegraphics[width=0.88\linewidth]{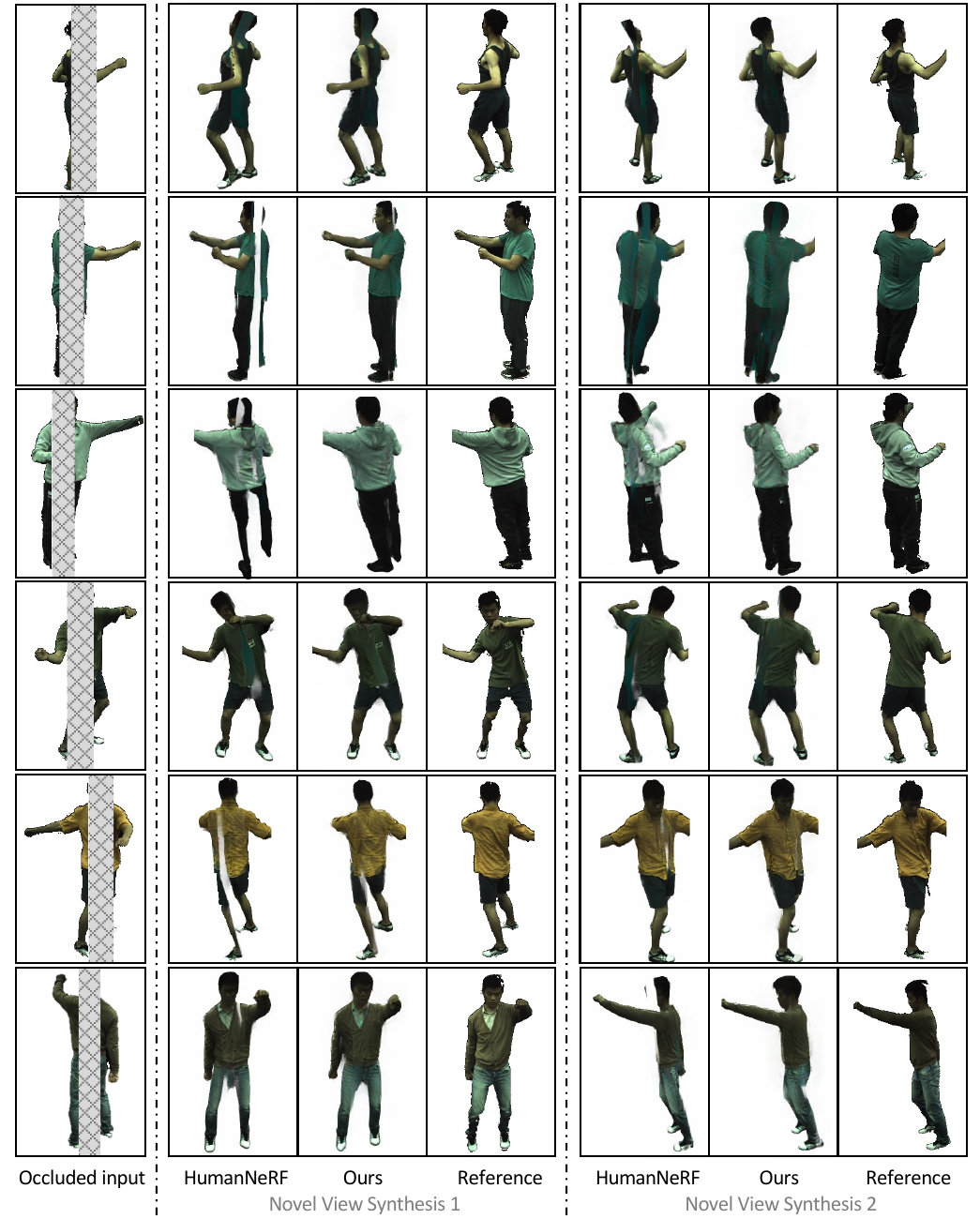}
    \caption{Qualitative results on \textbf{simulated occlusions} in the ZJU-MoCap dataset \cite{neuralbody}.}
    \label{fig:zju_result}
    \vspace{-1em}
\end{figure*}

\begin{figure*}[t]
    \centering
    \includegraphics[width=0.88\linewidth]{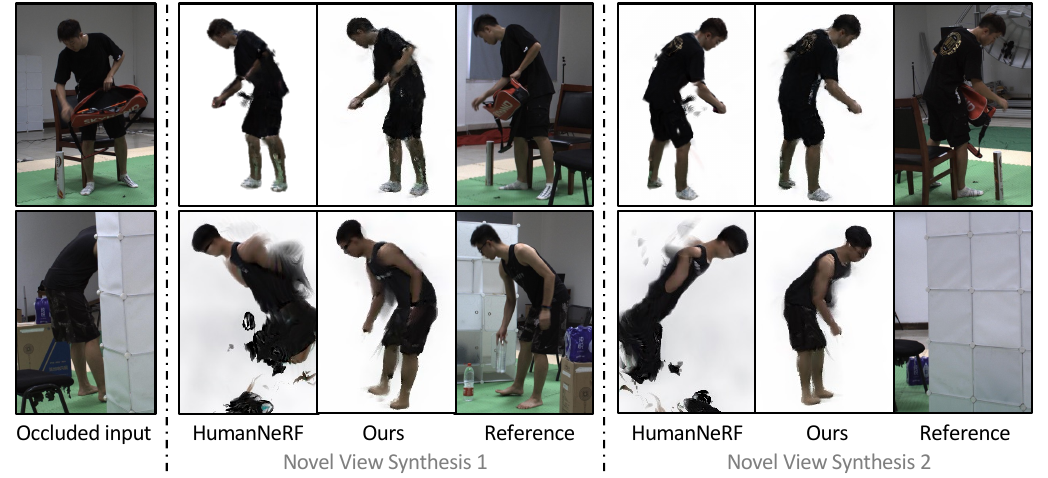}
    \caption{Qualitative results on \textbf{real-world} occlusions in OcMotion dataset \cite{huang2022occluded}.}
    \label{fig:ocmotion_result}
    \vspace{-1em}
\end{figure*}

\subsection{Comparison and Metrics}
We mainly compare our method with HumanNeRF \cite{weng2022humannerf}, the state-of-the-art human rendering method. We also compare against a baseline method Neural Body \cite{neuralbody} in supplementary materials. Note that all methods use identical prior information, including pre-computed binary human mask and SMPL/camera parameters. The extra visibility prior used in \methodname\ can be calculated from the videos.

Methods are compared qualitatively and quantitatively. For qualitative evaluations, we directly visualize novel views to assess the quality of the renderings. For quantitative evaluations, we rely on the commonly used peak signal-to-noise ratio (PSNR) and structural similarity (SSIM) metrics \cite{neuralbody, weng2022humannerf, peng2022selfnerf}. Previous methods computed these metrics on full-scale images, which contain a majority of transparent backgrounds. These regions are identical between predictions and references, which inflate the overall metrics. To focus on the quality of rendered humans, we compute metrics on the pixels with non-zero accumulated $\alpha$. For OcMotion, since there is no ground truth for real-world occlusions, we compute the metrics on the visible area only. We refer to the standard metrics as PSNR$_\text{full}$/SSIM$_\text{full}$ and modified metrics as PSNR$_\text{vis}$/SSIM$_\text{vis}$.

% Their good quantitative results are mainly attributed to the high metric values on the background

\subsection{Implementation Details} 
Using the loss formulated in \equationautorefname~\ref{eq:target}, we optimize \methodname\ with the Adam optimizer \cite{kingma2014adam}. We set the learning rate to $5\times10^{-4}$ for the regression MLP $\mathcal{F}$, $1\times10^{-4}$ for the parameterization vertices $\mathbf{v}$, and $5\times10^{-5}$ for the rest. $\lambda_1$, $\lambda_2$, and $\lambda_3$ were set to 0.2, 1.0, and 10.0 respectively. We adopted patch-wise sampling of rays, each with 128 sample points. Due to the usage of the hash grid, \methodname\ converges faster than HumanNeRF. As a result, we trained our models for only 10K iterations while HumanNeRF models for 40K iterations. 

\subsection{Results on Simulated Occlusions}

Qualitative comparison on ZJU-MoCap videos with simulated occlusions between HumanNeRF and  \methodname\ is shown in \figureautorefname~\ref{fig:zju_result}. \methodname\ is capable of rendering a mostly completed body geometry with sensible details filled in at occluded areas. On the contrary, HumanNeRF fails to recover occluded body parts and produces significant artifacts in the occluded areas. Additionally, the quantitative results in \tableautorefname~\ref{tab:main_results} show that \methodname\ surpasses HumanNeRF for all subjects and under both metrics by a great margin. Note that this straightforward simulation of occlusions is in fact uncommon in real-world settings, where obstacles should have various shapes and humans are able to move across the entire scene with interactions with obstacles. \emph{More comparisons against Neural Body \cite{neuralbody} can be found in supplementary materials.}

\subsection{Results on Real-world Occlusions}
For better validating on real-world scenes, we present the rendering results on OcMotion videos in \figureautorefname~\ref{fig:ocmotion_result}. For the video with mild occlusions (top row), \methodname\ outperforms HumanNeRF with a higher fidelity rendering of texture details and much fewer artifacts at non-human regions. For the video with severe occlusions (bottom row), \methodname\ is still able to generate novel views with high-level rendering quality. However, HumanNeRF fails completely in such challenging cases when most body parts are occluded. This validates the superiority of \methodname\ in real-world scenes. \methodname\ also exceeds HumanNeRF on quantitative benchmarks as indicated in \tableautorefname~\ref{tab:main_results}. Note that the metrics were computed on visible pixels in training images only, which ignored most of the artifacts HumanNeRF generated. \emph{More comparisons on real-world scenes can be found in supplementary materials.}

\begin{table*}[htbp]
\centering
\begin{tabular}{|c || c | c | c | c || c | c | c | c |}
\hline
\multirow{2}{*}{ZJU-MoCap} &  \multicolumn{4}{c||}{Subject \textbf{377}} & \multicolumn{4}{c|}{Subject \textbf{386}} \\ 
\cline{2-9}
 & PSNR$_\text{vis}$ & SSIM$_\text{vis}$ & PSNR$_\text{full}$ & SSIM$_\text{full}$ & PSNR$_\text{vis}$ & SSIM$_\text{vis}$ & PSNR$_\text{full}$ & SSIM$_\text{full}$ \\ 
\hline
HumanNeRF \cite{weng2022humannerf} &  11.29 & 0.5649 & 22.15 & 0.9612 &  9.491 & 0.4877 & 19.89 & 0.9531 \\
\hline
\methodname & \cellcolor{best_color}13.23  & \cellcolor{best_color}0.6097 & \cellcolor{best_color}23.43 & \cellcolor{best_color}0.9642 & \cellcolor{best_color}13.44 & \cellcolor{best_color}0.5974 & \cellcolor{best_color}23.66 & \cellcolor{best_color}0.9639\\
\hline
\hline
\multirow{2}{*}{ZJU-MoCap} &  \multicolumn{4}{c||}{Subject \textbf{387}} & \multicolumn{4}{c|}{Subject \textbf{392}} \\ 
\cline{2-9}
 & PSNR$_\text{vis}$ & SSIM$_\text{vis}$ & PSNR$_\text{full}$ & SSIM$_\text{full}$ & PSNR$_\text{vis}$ & SSIM$_\text{vis}$ & PSNR$_\text{full}$ & SSIM$_\text{full}$ \\ 
\hline
HumanNeRF \cite{weng2022humannerf} &  9.551 & 0.4140 & 19.47 & 0.9408 & 11.04 & 0.5290 & 21.01 & 0.9543 \\
\hline
\methodname & \cellcolor{best_color}13.27  & \cellcolor{best_color}0.5243 & \cellcolor{best_color}22.26 & \cellcolor{best_color}0.9513 & \cellcolor{best_color}13.00 & \cellcolor{best_color}0.5692 & \cellcolor{best_color}22.13 & \cellcolor{best_color}0.9575\\
\hline
\hline
\multirow{2}{*}{ZJU-MoCap} &  \multicolumn{4}{c||}{Subject \textbf{393}} & \multicolumn{4}{c|}{Subject \textbf{394}} \\ 
\cline{2-9}
 & PSNR$_\text{vis}$ & SSIM$_\text{vis}$ & PSNR$_\text{full}$ & SSIM$_\text{full}$ & PSNR$_\text{vis}$ & SSIM$_\text{vis}$ & PSNR$_\text{full}$ & SSIM$_\text{full}$ \\ 
\hline
HumanNeRF \cite{weng2022humannerf} &  10.86 & 0.4483 & 20.92 & 0.9476 & 10.55 & 0.4764 &  20.56 & 0.9489 \\
\hline
\methodname & \cellcolor{best_color}12.00  & \cellcolor{best_color}0.4655 & \cellcolor{best_color}21.58 & \cellcolor{best_color}0.9489 & \cellcolor{best_color}13.12 & \cellcolor{best_color}0.5317 & \cellcolor{best_color}22.06 & \cellcolor{best_color}0.9532\\
\hline
\hline
\multirow{2}{*}{OcMotion} &  \multicolumn{4}{c||}{Video \textbf{Mild}} & \multicolumn{4}{c|}{Video \textbf{Severe}} \\ 
\cline{2-9}
 & PSNR$_\text{vis}$ & SSIM$_\text{vis}$ & PSNR$_\text{full}$ & SSIM$_\text{full}$ & PSNR$_\text{vis}$ & SSIM$_\text{vis}$ & PSNR$_\text{full}$ & SSIM$_\text{full}$ \\ 
\hline
HumanNeRF \cite{weng2022humannerf} &  13.38 & 0.6544 & 21.18 & 0.9680 & 11.40 & 0.4545 & 17.96 & 0.9470 \\
\hline
\methodname & \cellcolor{best_color}14.56  & \cellcolor{best_color}0.6814 & \cellcolor{best_color}21.50 & \cellcolor{best_color}0.9695 & \cellcolor{best_color}14.95 & \cellcolor{best_color}0.5998 & \cellcolor{best_color}21.16 & \cellcolor{best_color}0.9692\\
\hline

% \hline
% \multirow{2}{*}{} &  \multicolumn{4}{c||}{ZJU-MoCap Subject \textbf{392}} & \multicolumn{4}{c|}{ZJU-MoCap Subject \textbf{393}}\\ 
% \cline{2-9}
%  &  PSNR$_\text{vis}$ & SSIM$_\text{vis}$ & PSNR$_\text{full}$ & SSIM$_\text{full}$ & PSNR$_\text{vis}$ & SSIM$_\text{vis}$ & PSNR$_\text{full}$ & SSIM$_\text{full}$  \\ 
% \hline
% HumanNeRF \cite{weng2022humannerf} & 30.10 & 0.9642 & 53.27 & \cellcolor{best_color}28.61 & 0.9590 & 59.05 & 29.10 & 0.959 \\
% \hline
% \methodname & \cellcolor{best_color}31.04 & \cellcolor{best_color}0.9705 & \cellcolor{best_color}32.12 & 28.31 & \cellcolor{best_color}0.9603 & \cellcolor{best_color}36.72 & \cellcolor{best_color}30.31 & \cellcolor{best_color}0.9642 \\
% \hline
% \hline
% \multirow{2}{*}{} &  \multicolumn{4}{c||}{OcMotion \textbf{Mild}} & \multicolumn{4}{c|}{OcMotion \textbf{Severe}}\\ 
% \cline{2-9}
%  &  PSNR$_\text{vis}$ & SSIM$_\text{vis}$ & PSNR$_\text{full}$ & SSIM$_\text{full}$ & PSNR$_\text{vis}$ & SSIM$_\text{vis}$ & PSNR$_\text{full}$ & SSIM$_\text{full}$  \\ 
% \hline
% HumanNeRF \cite{weng2022humannerf} & 30.10 & 0.9642 & 53.27 & \cellcolor{best_color}28.61 & 0.9590 & 59.05 & 29.10 & 0.959 \\
% \hline
% \methodname & \cellcolor{best_color}31.04 & \cellcolor{best_color}0.9705 & \cellcolor{best_color}32.12 & 28.31 & \cellcolor{best_color}0.9603 & \cellcolor{best_color}36.72 & \cellcolor{best_color}30.31 & \cellcolor{best_color}0.9642 \\
% \hline
\end{tabular}
\vspace{0.8em}
\caption{Quantitative comparison on the ZJU-MoCap and OcMotion datasets. We color cells that have the \colorbox{best_color}{best} metric values.}
\label{tab:main_results}
\vspace{-1em}
%\vspace{-14px}
\end{table*}

\subsection{Ablation Studies}

In this section, we conduct additional experiments by simply removing each of the proposed components from the \methodname\ framework to prove their effectiveness. Quantitative metrics are also presented in the figures. \emph{More ablation studies can be found in supplementary materials.}

\begin{figure}[h]
    \centering
    \includegraphics[width=\linewidth]{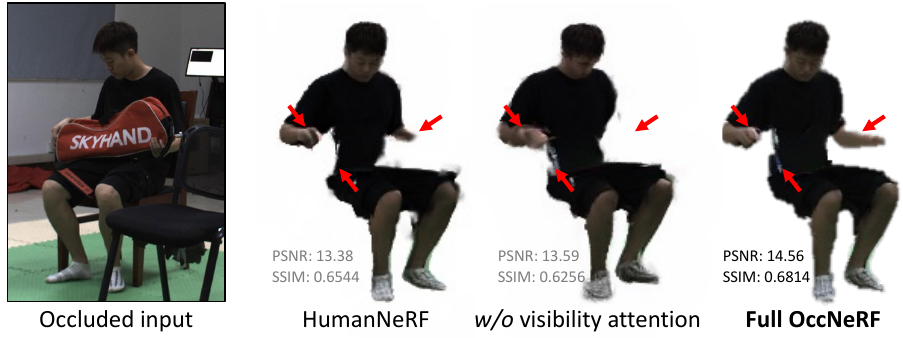}
    \caption{Our visibility attention improves rendering quality with more confident predictions at occluded areas with fewer blurs.}
    \label{fig:ablation3}
    
\end{figure}

% ablation3
\noindent\textbf{Impact of Visibility Attention.}  Our ablation studies start by proving the benefits of reformulating visibility priors as attention maps to be applied during surface-based rendering. \figureautorefname~\ref{fig:ablation3} shows that when disabling the attentive aggregation from \equationautorefname~\ref{eq:attentive_agg}, the model becomes less confident in occluded areas, resulting in more blurs.

\begin{figure}[t]
    \centering
    \includegraphics[width=\linewidth]{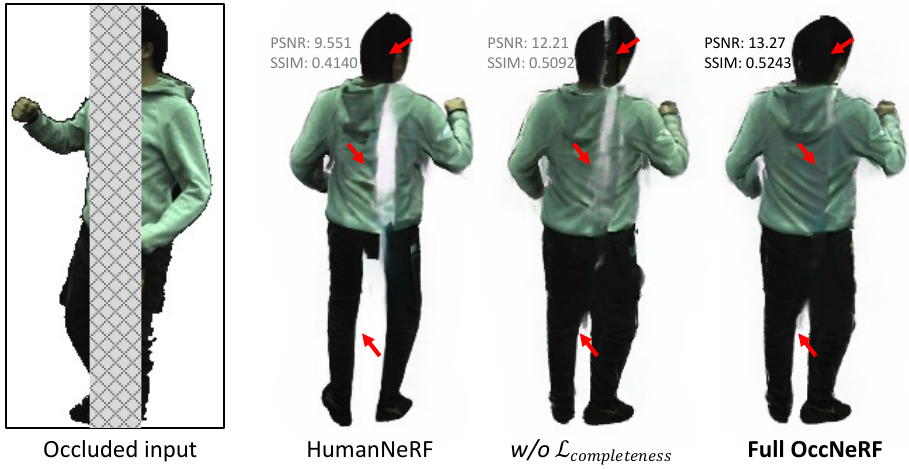}
    \caption{Our $\mathcal{L}_\text{comp}$ improves geometry completeness a step further when combined with the proposed rendering strategy.}
    \label{fig:ablation4}
    \vspace{-1em}
\end{figure}

% ablation4
\noindent\textbf{Impact of $\mathcal{L}_\text{comp}$.} The proposed completeness loss $\mathcal{L}_\text{comp}$ is designed to encourage high-density values at locations inside the SMPL mesh. When removing this loss, \figureautorefname~\ref{fig:ablation4} shows that our method cannot render a complete geometry anymore. However, with our surface-based rendering, we still yield better results than HumanNeRF.

% % ablation2 similar results but longer time
% \noindent\textbf{Impact of multi-scale representation.} As discussed in \sectionautorefname~\ref{}, representing parameterization vertices at multiple scales covers a wide range of the body surface more efficiently. Under the same computation burden, we managed to search for the nearest vertices at the finest scale only to validate the effectiveness of multi-scale representation. The result $\mathbf{X}$ shows that XXXX.

\begin{figure}[t]
    \centering
    \includegraphics[width=\linewidth]{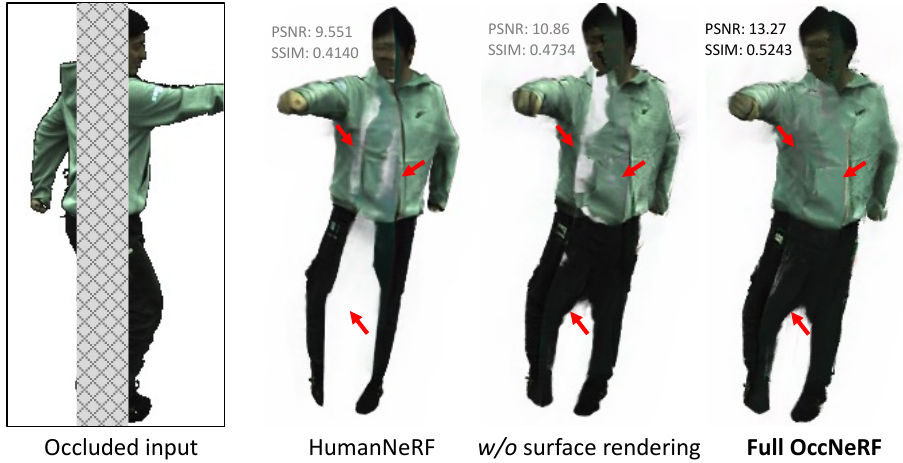}
    \caption{Our surface-based rendering method fills in the occluded parts with both accurate geometry and appropriate appearance.}
    \label{fig:ablation1}
    \vspace{-1.5em}
\end{figure}

% ablation1
\noindent\textbf{Impact of Surface-based Rendering.} As discussed in \sectionautorefname~\ref{sec:3.2}, we claimed that our proposed rendering strategy enables $\mathcal{F}(\cdot)$ to condition on inputs better with more overlaps. Here we validate the necessity of such a design by removing it from the framework. We, however, still keep the hash grid encoding to see its impact. According to \figureautorefname~\ref{fig:ablation1}, the hash grid encoding alone is not able to achieve comparable performance to our full \methodname. It has to be equipped together with the proposed rendering strategy. This validates that major performance improvements do come from surface-based rendering.

\section{Discussions and Conclusion}

\noindent\textbf{Discussions.} It is difficult to optimize scene-specific neural radiance fields under occlusions. There is neither a ground truth for the occluded parts nor external information from different scenes to inpaint the missing area. \methodname\ achieves rendering of the occluded regions by referring to nearby visible correspondences and enforcing complete geometry. However, \methodname\ can yield subtle artifacts. This is because we have more parameters to optimize and fewer training data due to occlusions. Since no external information is accessible, \methodname\ is not capable of inpainting an area that has never been seen in the video. The above limitations can be overcome with a better geometry prior \cite{jiang2022selfrecon} and a cross-scene training strategy \cite{chen2022gpnerf}. Although the hash grid encoding accelerates the convergence at training, \methodname\ runs relatively slower than HumanNeRF at inference.

\noindent\textbf{Conclusion.} We proposed \methodname\ for rendering humans from object-occluded monocular videos. Most existing methods assume clear views of the entire human body without any interference, which is not feasible in real-world scenes. We designed a surface-based rendering strategy that incorporates geometry and visibility priors to assist rendering under occlusions. Moreover, our novel loss function is also able to help maintain geometry completeness. In our experiments, we compared \methodname\ against the state-of-the-art method under both simulated and real-world video occlusions. Our state-of-the-art results set up a new benchmark in this field of research.

\vspace{-10pt}
\paragraph{Acknowledgments.} This work was partially funded by the Gordon and Betty Moore Foundation, Panasonic Holdings Corporation, NSF RI \#2211258, and Stanford HAI.

%-------------------------------------------------------------------------

{\small
\bibliographystyle{ieee_fullname}
\bibliography{egbib}

\begin{thebibliography}{10}\itemsep=-1pt

\bibitem{alldieck2018video}
Thiemo Alldieck, Marcus Magnor, Weipeng Xu, Christian Theobalt, and Gerard
  Pons-Moll.
\newblock Video based reconstruction of 3d people models.
\newblock In {\em Proceedings of the IEEE Conference on Computer Vision and
  Pattern Recognition}, pages 8387--8397, 2018.

\bibitem{barron2021mip}
Jonathan~T Barron, Ben Mildenhall, Matthew Tancik, Peter Hedman, Ricardo
  Martin-Brualla, and Pratul~P Srinivasan.
\newblock Mip-nerf: A multiscale representation for anti-aliasing neural
  radiance fields.
\newblock In {\em Proceedings of the IEEE/CVF International Conference on
  Computer Vision}, pages 5855--5864, 2021.

\bibitem{barron2022mip}
Jonathan~T Barron, Ben Mildenhall, Dor Verbin, Pratul~P Srinivasan, and Peter
  Hedman.
\newblock Mip-nerf 360: Unbounded anti-aliased neural radiance fields.
\newblock In {\em Proceedings of the IEEE/CVF Conference on Computer Vision and
  Pattern Recognition}, pages 5470--5479, 2022.

\bibitem{bergman2022generative}
Alexander~W Bergman, Petr Kellnhofer, Yifan Wang, Eric~R Chan, David~B Lindell,
  and Gordon Wetzstein.
\newblock Generative neural articulated radiance fields.
\newblock {\em arXiv preprint arXiv:2206.14314}, 2022.

\bibitem{biggs20203d}
Benjamin Biggs, David Novotny, Sebastien Ehrhardt, Hanbyul Joo, Ben Graham, and
  Andrea Vedaldi.
\newblock 3d multi-bodies: Fitting sets of plausible 3d human models to
  ambiguous image data.
\newblock {\em Advances in Neural Information Processing Systems},
  33:20496--20507, 2020.

\bibitem{Bogo:ECCV:2016}
Federica Bogo, Angjoo Kanazawa, Christoph Lassner, Peter Gehler, Javier Romero,
  and Michael~J. Black.
\newblock Keep it {SMPL}: Automatic estimation of {3D} human pose and shape
  from a single image.
\newblock In {\em Computer Vision -- ECCV 2016}, Lecture Notes in Computer
  Science. Springer International Publishing, Oct. 2016.

\bibitem{chen2021animatable}
Jianchuan Chen, Ying Zhang, Di Kang, Xuefei Zhe, Linchao Bao, Xu Jia, and
  Huchuan Lu.
\newblock Animatable neural radiance fields from monocular rgb videos.
\newblock {\em arXiv preprint arXiv:2106.13629}, 2021.

\bibitem{chen2022gpnerf}
Mingfei Chen, Jianfeng Zhang, Xiangyu Xu, Lijuan Liu, Yujun Cai, Jiashi Feng,
  and Shuicheng Yan.
\newblock Geometry-guided progressive nerf for generalizable and efficient
  neural human rendering.
\newblock In {\em ECCV}, 2022.

\bibitem{chen2022gdna}
Xu Chen, Tianjian Jiang, Jie Song, Jinlong Yang, Michael~J Black, Andreas
  Geiger, and Otmar Hilliges.
\newblock gdna: Towards generative detailed neural avatars.
\newblock In {\em Proceedings of the IEEE/CVF Conference on Computer Vision and
  Pattern Recognition}, pages 20427--20437, 2022.

\bibitem{chen2022hallucinated}
Xingyu Chen, Qi Zhang, Xiaoyu Li, Yue Chen, Ying Feng, Xuan Wang, and Jue Wang.
\newblock Hallucinated neural radiance fields in the wild.
\newblock In {\em Proceedings of the IEEE/CVF Conference on Computer Vision and
  Pattern Recognition}, pages 12943--12952, 2022.

\bibitem{chibane2020implicit}
Julian Chibane, Thiemo Alldieck, and Gerard Pons-Moll.
\newblock Implicit functions in feature space for 3d shape reconstruction and
  completion.
\newblock In {\em Proceedings of the IEEE/CVF conference on computer vision and
  pattern recognition}, pages 6970--6981, 2020.

\bibitem{collet2015high}
Alvaro Collet, Ming Chuang, Pat Sweeney, Don Gillett, Dennis Evseev, David
  Calabrese, Hugues Hoppe, Adam Kirk, and Steve Sullivan.
\newblock High-quality streamable free-viewpoint video.
\newblock {\em ACM Transactions on Graphics (ToG)}, 34(4):1--13, 2015.

\bibitem{corona2021smplicit}
Enric Corona, Albert Pumarola, Guillem Alenya, Gerard Pons-Moll, and Francesc
  Moreno-Noguer.
\newblock Smplicit: Topology-aware generative model for clothed people.
\newblock In {\em Proceedings of the IEEE/CVF conference on computer vision and
  pattern recognition}, pages 11875--11885, 2021.

\bibitem{deng2020nasa}
Boyang Deng, John~P Lewis, Timothy Jeruzalski, Gerard Pons-Moll, Geoffrey
  Hinton, Mohammad Norouzi, and Andrea Tagliasacchi.
\newblock Nasa neural articulated shape approximation.
\newblock In {\em Computer Vision--ECCV 2020: 16th European Conference,
  Glasgow, UK, August 23--28, 2020, Proceedings, Part VII 16}, pages 612--628.
  Springer, 2020.

\bibitem{dou2016fusion4d}
Mingsong Dou, Sameh Khamis, Yury Degtyarev, Philip Davidson, Sean~Ryan Fanello,
  Adarsh Kowdle, Sergio~Orts Escolano, Christoph Rhemann, David Kim, Jonathan
  Taylor, et~al.
\newblock Fusion4d: Real-time performance capture of challenging scenes.
\newblock {\em ACM Transactions on Graphics (ToG)}, 35(4):1--13, 2016.

\bibitem{endo2022gaitforemer}
Mark Endo, Kathleen~L Poston, Edith~V Sullivan, Li Fei-Fei, Kilian~M Pohl, and
  Ehsan Adeli.
\newblock Gaitforemer: Self-supervised pre-training of transformers via human
  motion forecasting for few-shot gait impairment severity estimation.
\newblock In {\em Medical Image Computing and Computer Assisted
  Intervention--MICCAI 2022: 25th International Conference, Singapore,
  September 18--22, 2022, Proceedings, Part VIII}, pages 130--139. Springer,
  2022.

\bibitem{feng2022capturing}
Yao Feng, Jinlong Yang, Marc Pollefeys, Michael~J Black, and Timo Bolkart.
\newblock Capturing and animation of body and clothing from monocular video.
\newblock {\em arXiv preprint arXiv:2210.01868}, 2022.

\bibitem{gao2021dynamic}
Chen Gao, Ayush Saraf, Johannes Kopf, and Jia-Bin Huang.
\newblock Dynamic view synthesis from dynamic monocular video.
\newblock In {\em Proceedings of the IEEE/CVF International Conference on
  Computer Vision}, pages 5712--5721, 2021.

\bibitem{gerats20223d}
Beerend~GA Gerats, Jelmer~M Wolterink, and Ivo~AMJ Broeders.
\newblock 3d human pose estimation in multi-view operating room videos using
  differentiable camera projections.
\newblock {\em Computer Methods in Biomechanics and Biomedical Engineering:
  Imaging \& Visualization}, pages 1--9, 2022.

\bibitem{he2020geo}
Tong He, John Collomosse, Hailin Jin, and Stefano Soatto.
\newblock Geo-pifu: Geometry and pixel aligned implicit functions for
  single-view human reconstruction.
\newblock {\em Advances in Neural Information Processing Systems},
  33:9276--9287, 2020.

\bibitem{he2021arch++}
Tong He, Yuanlu Xu, Shunsuke Saito, Stefano Soatto, and Tony Tung.
\newblock Arch++: Animation-ready clothed human reconstruction revisited.
\newblock In {\em Proceedings of the IEEE/CVF international conference on
  computer vision}, pages 11046--11056, 2021.

\bibitem{hedman2021baking}
Peter Hedman, Pratul~P Srinivasan, Ben Mildenhall, Jonathan~T Barron, and Paul
  Debevec.
\newblock Baking neural radiance fields for real-time view synthesis.
\newblock In {\em Proceedings of the IEEE/CVF International Conference on
  Computer Vision}, pages 5875--5884, 2021.

\bibitem{hong2021stereopifu}
Yang Hong, Juyong Zhang, Boyi Jiang, Yudong Guo, Ligang Liu, and Hujun Bao.
\newblock Stereopifu: Depth aware clothed human digitization via stereo vision.
\newblock In {\em Proceedings of the IEEE/CVF Conference on Computer Vision and
  Pattern Recognition}, pages 535--545, 2021.

\bibitem{huang2022occluded}
Buzhen Huang, Yuan Shu, Jingyi Ju, and Yangang Wang.
\newblock Occluded human body capture with self-supervised spatial-temporal
  motion prior.
\newblock {\em arXiv preprint arXiv:2207.05375}, 2022.

\bibitem{huang2021dynamic}
Buzhen Huang, Yuan Shu, Tianshu Zhang, and Yangang Wang.
\newblock Dynamic multi-person mesh recovery from uncalibrated multi-view
  cameras.
\newblock In {\em 2021 International Conference on 3D Vision (3DV)}, pages
  710--720. IEEE, 2021.

\bibitem{huang2020arch}
Zeng Huang, Yuanlu Xu, Christoph Lassner, Hao Li, and Tony Tung.
\newblock Arch: Animatable reconstruction of clothed humans.
\newblock In {\em Proceedings of the IEEE/CVF Conference on Computer Vision and
  Pattern Recognition}, pages 3093--3102, 2020.

\bibitem{jiang2022selfrecon}
Boyi Jiang, Yang Hong, Hujun Bao, and Juyong Zhang.
\newblock Selfrecon: Self reconstruction your digital avatar from monocular
  video.
\newblock In {\em Proceedings of the IEEE/CVF Conference on Computer Vision and
  Pattern Recognition}, pages 5605--5615, 2022.

\bibitem{jiang2022instantavatar}
Tianjian Jiang, Xu Chen, Jie Song, and Otmar Hilliges.
\newblock Instantavatar: Learning avatars from monocular video in 60 seconds.
\newblock {\em arXiv preprint arXiv:2212.10550}, 2022.

\bibitem{jiang2020coherent}
Wen Jiang, Nikos Kolotouros, Georgios Pavlakos, Xiaowei Zhou, and Kostas
  Daniilidis.
\newblock Coherent reconstruction of multiple humans from a single image.
\newblock In {\em Proceedings of the IEEE/CVF conference on computer vision and
  pattern recognition}, pages 5579--5588, 2020.

\bibitem{jiang2022neuman}
Wei Jiang, Kwang~Moo Yi, Golnoosh Samei, Oncel Tuzel, and Anurag Ranjan.
\newblock Neuman: Neural human radiance field from a single video.
\newblock In {\em Computer Vision--ECCV 2022: 17th European Conference, Tel
  Aviv, Israel, October 23--27, 2022, Proceedings, Part XXXII}, pages 402--418.
  Springer, 2022.

\bibitem{humanMotionKZFM19}
Angjoo Kanazawa, Jason~Y. Zhang, Panna Felsen, and Jitendra Malik.
\newblock Learning 3d human dynamics from video.
\newblock In {\em Computer Vision and Pattern Recognition (CVPR)}, 2019.

\bibitem{kingma2014adam}
Diederik~P Kingma and Jimmy Ba.
\newblock Adam: A method for stochastic optimization.
\newblock {\em arXiv preprint arXiv:1412.6980}, 2014.

\bibitem{kocabas2019vibe}
Muhammed Kocabas, Nikos Athanasiou, and Michael~J. Black.
\newblock Vibe: Video inference for human body pose and shape estimation.
\newblock In {\em The IEEE Conference on Computer Vision and Pattern
  Recognition (CVPR)}, June 2020.

\bibitem{kocabas2021pare}
Muhammed Kocabas, Chun-Hao~P Huang, Otmar Hilliges, and Michael~J Black.
\newblock Pare: Part attention regressor for 3d human body estimation.
\newblock In {\em Proceedings of the IEEE/CVF International Conference on
  Computer Vision}, pages 11127--11137, 2021.

\bibitem{li2022tava}
Ruilong Li, Julian Tanke, Minh Vo, Michael Zollh{\"o}fer, J{\"u}rgen Gall,
  Angjoo Kanazawa, and Christoph Lassner.
\newblock Tava: Template-free animatable volumetric actors.
\newblock In {\em Computer Vision--ECCV 2022: 17th European Conference, Tel
  Aviv, Israel, October 23--27, 2022, Proceedings, Part XXXII}, pages 419--436.
  Springer, 2022.

\bibitem{li2021neural}
Zhengqi Li, Simon Niklaus, Noah Snavely, and Oliver Wang.
\newblock Neural scene flow fields for space-time view synthesis of dynamic
  scenes.
\newblock In {\em Proceedings of the IEEE/CVF Conference on Computer Vision and
  Pattern Recognition}, pages 6498--6508, 2021.

\bibitem{liu2021neural}
Lingjie Liu, Marc Habermann, Viktor Rudnev, Kripasindhu Sarkar, Jiatao Gu, and
  Christian Theobalt.
\newblock Neural actor: Neural free-view synthesis of human actors with pose
  control.
\newblock {\em ACM Transactions on Graphics (TOG)}, 40(6):1--16, 2021.

\bibitem{liu2022neural}
Yuan Liu, Sida Peng, Lingjie Liu, Qianqian Wang, Peng Wang, Christian Theobalt,
  Xiaowei Zhou, and Wenping Wang.
\newblock Neural rays for occlusion-aware image-based rendering.
\newblock In {\em Proceedings of the IEEE/CVF Conference on Computer Vision and
  Pattern Recognition}, pages 7824--7833, 2022.

\bibitem{lombardi2019neural}
Stephen Lombardi, Tomas Simon, Jason Saragih, Gabriel Schwartz, Andreas
  Lehrmann, and Yaser Sheikh.
\newblock Neural volumes: Learning dynamic renderable volumes from images.
\newblock {\em arXiv preprint arXiv:1906.07751}, 2019.

\bibitem{SMPL}
Matthew Loper, Naureen Mahmood, Javier Romero, Gerard Pons-Moll, and Michael~J.
  Black.
\newblock {SMPL}: A skinned multi-person linear model.
\newblock {\em ACM Trans. Graphics (Proc. SIGGRAPH Asia)}, 34(6):248:1--248:16,
  Oct. 2015.

\bibitem{Loubet2019Reparameterizing}
Guillaume Loubet, Nicolas Holzschuch, and Wenzel Jakob.
\newblock Reparameterizing discontinuous integrands for differentiable
  rendering.
\newblock {\em Transactions on Graphics (Proceedings of SIGGRAPH Asia)}, 38(6),
  Dec. 2019.

\bibitem{matusik2000image}
Wojciech Matusik, Chris Buehler, Ramesh Raskar, Steven~J Gortler, and Leonard
  McMillan.
\newblock Image-based visual hulls.
\newblock In {\em Proceedings of the 27th annual conference on Computer
  graphics and interactive techniques}, pages 369--374, 2000.

\bibitem{nerf}
Ben Mildenhall, Pratul~P Srinivasan, Matthew Tancik, Jonathan~T Barron, Ravi
  Ramamoorthi, and Ren Ng.
\newblock Nerf: Representing scenes as neural radiance fields for view
  synthesis.
\newblock {\em Communications of the ACM}, 65(1):99--106, 2021.

\bibitem{muller2022instant}
Thomas M{\"u}ller, Alex Evans, Christoph Schied, and Alexander Keller.
\newblock Instant neural graphics primitives with a multiresolution hash
  encoding.
\newblock {\em arXiv preprint arXiv:2201.05989}, 2022.

\bibitem{noguchi2021neural}
Atsuhiro Noguchi, Xiao Sun, Stephen Lin, and Tatsuya Harada.
\newblock Neural articulated radiance field.
\newblock In {\em Proceedings of the IEEE/CVF International Conference on
  Computer Vision}, pages 5762--5772, 2021.

\bibitem{ost2021neural}
Julian Ost, Fahim Mannan, Nils Thuerey, Julian Knodt, and Felix Heide.
\newblock Neural scene graphs for dynamic scenes.
\newblock In {\em Proceedings of the IEEE/CVF Conference on Computer Vision and
  Pattern Recognition}, pages 2856--2865, 2021.

\bibitem{park2021nerfies}
Keunhong Park, Utkarsh Sinha, Jonathan~T Barron, Sofien Bouaziz, Dan~B Goldman,
  Steven~M Seitz, and Ricardo Martin-Brualla.
\newblock Nerfies: Deformable neural radiance fields.
\newblock In {\em Proceedings of the IEEE/CVF International Conference on
  Computer Vision}, pages 5865--5874, 2021.

\bibitem{park2021hypernerf}
Keunhong Park, Utkarsh Sinha, Peter Hedman, Jonathan~T Barron, Sofien Bouaziz,
  Dan~B Goldman, Ricardo Martin-Brualla, and Steven~M Seitz.
\newblock Hypernerf: A higher-dimensional representation for topologically
  varying neural radiance fields.
\newblock {\em arXiv preprint arXiv:2106.13228}, 2021.

\bibitem{peng2022selfnerf}
Bo Peng, Jun Hu, Jingtao Zhou, and Juyong Zhang.
\newblock Selfnerf: Fast training nerf for human from monocular self-rotating
  video.
\newblock {\em arXiv preprint arXiv:2210.01651}, 2022.

\bibitem{peng2021animatable}
Sida Peng, Junting Dong, Qianqian Wang, Shangzhan Zhang, Qing Shuai, Xiaowei
  Zhou, and Hujun Bao.
\newblock Animatable neural radiance fields for modeling dynamic human bodies.
\newblock In {\em Proceedings of the IEEE/CVF International Conference on
  Computer Vision}, pages 14314--14323, 2021.

\bibitem{neuralbody}
Sida Peng, Yuanqing Zhang, Yinghao Xu, Qianqian Wang, Qing Shuai, Hujun Bao,
  and Xiaowei Zhou.
\newblock Neural body: Implicit neural representations with structured latent
  codes for novel view synthesis of dynamic humans.
\newblock In {\em Proceedings of the IEEE/CVF Conference on Computer Vision and
  Pattern Recognition}, pages 9054--9063, 2021.

\bibitem{pumarola2021d}
Albert Pumarola, Enric Corona, Gerard Pons-Moll, and Francesc Moreno-Noguer.
\newblock D-nerf: Neural radiance fields for dynamic scenes.
\newblock In {\em Proceedings of the IEEE/CVF Conference on Computer Vision and
  Pattern Recognition}, pages 10318--10327, 2021.

\bibitem{qi2017pointnet}
Charles~R Qi, Hao Su, Kaichun Mo, and Leonidas~J Guibas.
\newblock Pointnet: Deep learning on point sets for 3d classification and
  segmentation.
\newblock In {\em Proceedings of the IEEE conference on computer vision and
  pattern recognition}, pages 652--660, 2017.

\bibitem{rockwell2020full}
Chris Rockwell and David~F Fouhey.
\newblock Full-body awareness from partial observations.
\newblock In {\em Computer Vision--ECCV 2020: 16th European Conference,
  Glasgow, UK, August 23--28, 2020, Proceedings, Part XVII 16}, pages 522--539.
  Springer, 2020.

\bibitem{saito2019pifu}
Shunsuke Saito, Zeng Huang, Ryota Natsume, Shigeo Morishima, Angjoo Kanazawa,
  and Hao Li.
\newblock Pifu: Pixel-aligned implicit function for high-resolution clothed
  human digitization.
\newblock In {\em Proceedings of the IEEE/CVF international conference on
  computer vision}, pages 2304--2314, 2019.

\bibitem{sarandi2018robust}
Istv{\'a}n S{\'a}r{\'a}ndi, Timm Linder, Kai~O Arras, and Bastian Leibe.
\newblock How robust is 3d human pose estimation to occlusion?
\newblock {\em arXiv preprint arXiv:1808.09316}, 2018.

\bibitem{srinivasan2021nerv}
Pratul~P Srinivasan, Boyang Deng, Xiuming Zhang, Matthew Tancik, Ben
  Mildenhall, and Jonathan~T Barron.
\newblock Nerv: Neural reflectance and visibility fields for relighting and
  view synthesis.
\newblock In {\em Proceedings of the IEEE/CVF Conference on Computer Vision and
  Pattern Recognition}, pages 7495--7504, 2021.

\bibitem{sun2021direct}
Cheng Sun, Min Sun, and Hwann-Tzong Chen.
\newblock Direct voxel grid optimization: Super-fast convergence for radiance
  fields reconstruction.
\newblock {\em CVPR}, 2022.

\bibitem{sun2021monocular}
Yu Sun, Qian Bao, Wu Liu, Yili Fu, Michael~J Black, and Tao Mei.
\newblock Monocular, one-stage, regression of multiple 3d people.
\newblock In {\em Proceedings of the IEEE/CVF international conference on
  computer vision}, pages 11179--11188, 2021.

\bibitem{BEV}
Yu Sun, Wu Liu, Qian Bao, Yili Fu, Tao Mei, and Michael~J Black.
\newblock Putting people in their place: Monocular regression of 3d people in
  depth.
\newblock In {\em CVPR}, 2022.

\bibitem{tancik2020fourier}
Matthew Tancik, Pratul Srinivasan, Ben Mildenhall, Sara Fridovich-Keil, Nithin
  Raghavan, Utkarsh Singhal, Ravi Ramamoorthi, Jonathan Barron, and Ren Ng.
\newblock Fourier features let networks learn high frequency functions in low
  dimensional domains.
\newblock {\em Advances in Neural Information Processing Systems},
  33:7537--7547, 2020.

\bibitem{tiwari2021neural}
Garvita Tiwari, Nikolaos Sarafianos, Tony Tung, and Gerard Pons-Moll.
\newblock Neural-gif: Neural generalized implicit functions for animating
  people in clothing.
\newblock In {\em Proceedings of the IEEE/CVF International Conference on
  Computer Vision}, pages 11708--11718, 2021.

\bibitem{verbin2022ref}
Dor Verbin, Peter Hedman, Ben Mildenhall, Todd Zickler, Jonathan~T Barron, and
  Pratul~P Srinivasan.
\newblock Ref-nerf: Structured view-dependent appearance for neural radiance
  fields.
\newblock In {\em 2022 IEEE/CVF Conference on Computer Vision and Pattern
  Recognition (CVPR)}, pages 5481--5490. IEEE, 2022.

\bibitem{wang2022arah}
Shaofei Wang, Katja Schwarz, Andreas Geiger, and Siyu Tang.
\newblock Arah: Animatable volume rendering of articulated human sdfs.
\newblock In {\em Computer Vision--ECCV 2022: 17th European Conference, Tel
  Aviv, Israel, October 23--27, 2022, Proceedings, Part XXXII}, pages 1--19.
  Springer, 2022.

\bibitem{weng2020vid2actor}
Chung-Yi Weng, Brian Curless, and Ira Kemelmacher-Shlizerman.
\newblock Vid2actor: Free-viewpoint animatable person synthesis from video in
  the wild.
\newblock {\em arXiv preprint arXiv:2012.12884}, 2020.

\bibitem{weng2022humannerf}
Chung-Yi Weng, Brian Curless, Pratul~P Srinivasan, Jonathan~T Barron, and Ira
  Kemelmacher-Shlizerman.
\newblock Humannerf: Free-viewpoint rendering of moving people from monocular
  video.
\newblock In {\em Proceedings of the IEEE/CVF Conference on Computer Vision and
  Pattern Recognition}, pages 16210--16220, 2022.

\bibitem{weng2023personnerf}
Chung-Yi Weng, Pratul~P Srinivasan, Brian Curless, and Ira
  Kemelmacher-Shlizerman.
\newblock Personnerf: Personalized reconstruction from photo collections.
\newblock {\em arXiv preprint arXiv:2302.08504}, 2023.

\bibitem{xu2021h}
Hongyi Xu, Thiemo Alldieck, and Cristian Sminchisescu.
\newblock H-nerf: Neural radiance fields for rendering and temporal
  reconstruction of humans in motion.
\newblock {\em Advances in Neural Information Processing Systems},
  34:14955--14966, 2021.

\bibitem{yang2022lasor}
Kaibing Yang, Renshu Gu, Maoyu Wang, Masahiro Toyoura, and Gang Xu.
\newblock Lasor: Learning accurate 3d human pose and shape via synthetic
  occlusion-aware data and neural mesh rendering.
\newblock {\em IEEE Transactions on Image Processing}, 31:1938--1948, 2022.

\bibitem{yang2021s3}
Ze Yang, Shenlong Wang, Sivabalan Manivasagam, Zeng Huang, Wei-Chiu Ma, Xinchen
  Yan, Ersin Yumer, and Raquel Urtasun.
\newblock S3: Neural shape, skeleton, and skinning fields for 3d human
  modeling.
\newblock In {\em Proceedings of the IEEE/CVF conference on computer vision and
  pattern recognition}, pages 13284--13293, 2021.

\bibitem{zhang2021body}
Jianfeng Zhang, Dongdong Yu, Jun~Hao Liew, Xuecheng Nie, and Jiashi Feng.
\newblock Body meshes as points.
\newblock In {\em Proceedings of the IEEE/CVF Conference on Computer Vision and
  Pattern Recognition}, pages 546--556, 2021.

\bibitem{zhang2018unreasonable}
Richard Zhang, Phillip Isola, Alexei~A Efros, Eli Shechtman, and Oliver Wang.
\newblock The unreasonable effectiveness of deep features as a perceptual
  metric.
\newblock In {\em Proceedings of the IEEE conference on computer vision and
  pattern recognition}, pages 586--595, 2018.

\bibitem{zhang2020object}
Tianshu Zhang, Buzhen Huang, and Yangang Wang.
\newblock Object-occluded human shape and pose estimation from a single color
  image.
\newblock In {\em Proceedings of the IEEE/CVF conference on computer vision and
  pattern recognition}, pages 7376--7385, 2020.

\end{thebibliography}
}

\end{document}